\documentclass[a4paper, 10 pt, conference]{ieeeconf}  

\IEEEoverridecommandlockouts                              

\usepackage{cite}
\usepackage{comment}
\usepackage{amsmath,amssymb,amsfonts}
\usepackage{algorithm}
\usepackage{algpseudocode}
\usepackage{graphicx}
\usepackage{textcomp}
\usepackage{stfloats}
\usepackage{xcolor}
\usepackage[T1]{fontenc} 

\usepackage{hyperref}
\usepackage{tabularx}
\usepackage{subcaption}
\usepackage{caption}
\usepackage{multirow}
\usepackage{epstopdf}
\usepackage{epsfig}
\usepackage{float}
\usepackage{units}
\usepackage{booktabs} 
\graphicspath{ {Images/} }
\DeclareGraphicsExtensions{.pdf,.eps}

\def\BibTeX{{\rm B\kern-.05em{\sc i\kern-.025em b}\kern-.08em
    T\kern-.1667em\lower.7ex\hbox{E}\kern-.125emX}}
\begin{document}
\title{Multimodal Dataset from Harsh Sub-Terranean Environment with Aerosol Particles for Frontier Exploration}
\author{Alexander Kyuroson$^{1}$, Niklas Dahlquist$^{1}$, Nikolaos Stathoulopoulos$^{1}$, Vignesh Kottayam Viswanathan$^{1}$, \\Anton Koval$^{1}$ and George Nikolakopoulos$^{1}$%
\thanks{This work has been funded by the European Unions Horizon 2020 Research and Innovation Programme under the Grant Agreement No. 101003591 NEX-GEN SIMS.}
\thanks{$^{1}$ The authors are with the Robotics and AI Team, Department of Computer, Electrical and Space Engineering, Lule\aa\,\, University of Technology, Lule\aa\,\,}
\thanks{Corresponding Author's email: {\tt\small akyuroson@gmail.com}}
}
\maketitle

\begin{abstract}

Algorithms for autonomous navigation in environments without Global Navigation Satellite System (GNSS) coverage mainly rely on onboard perception systems. These systems commonly incorporate sensors like cameras and Light Detection and Rangings (LiDARs), the performance of which may degrade in the presence of aerosol particles. Thus, there is a need of fusing acquired data from these sensors with data from Radio Detection and Rangings (RADARs) which can penetrate through such particles. Overall, this will improve the performance of localization and collision avoidance algorithms under such environmental conditions. This paper introduces a multimodal dataset from the harsh and unstructured underground environment with aerosol particles. A detailed description of the onboard sensors and the environment, where the dataset is collected are presented to enable full evaluation of acquired data. Furthermore, the dataset contains synchronized raw data measurements from all onboard sensors in Robot Operating System (ROS) format to facilitate the evaluation of navigation, and localization algorithms in such environments. In contrast to the existing datasets, the focus of this paper is not only to capture both temporal and spatial data diversities but also to present the impact of harsh conditions on captured data. Therefore, to validate the dataset, a preliminary comparison of odometry from onboard LiDARs is presented.

\end{abstract}

\section{Introduction} \label{intro}

Recent advancements in the field of autonomous robots navigation in the Sub-Terranean (SubT) environments such as mining tunnels~\cite{rogers2020test, tunnelcircuit}, urban SubT areas~\cite{rogers2021darpa}, and natural caves~\cite{koval2020subterranean} have dramatically increased the demand for the deployment of robotic platforms for various missions. In these GNSS-denied, harsh, and unstructured environments~\cite{mansouri2020nmpcmine, mansouri2020deploying}, Simultaneous Localization and Mapping (SLAM) is solely reliant on onboard sensors to allow these platforms to navigate and detect obstacles~\cite{caesar2020nuscenes} and to ensure operational safety~\cite{rogers2020test, giubilato2022slam}. To accomplish any given task while navigating in the extremely hazardous SubT environment, a combination of RADAR, LiDAR, and vision-based sensor suites is required for robust pose estimation and mapping. Existing SLAM implementations leverage various methods such as Visual-inertial~\cite{leutenegger2015keyframe}, LiDAR-inertial~\cite{hess2016real}, or RADAR-inertial~\cite{lu2020milliego, mostajabi2020high}. However, the quality of the estimated odometry from the aforementioned sensors which are commonly used in a variety of scenarios in the field of robotics is severely obstructed due to Visually Degrading Conditions (VDCs) such as darkness~\cite{kasper2019benchmark} and aerosol particles in form of smoke~\cite{lu2020see}, dust~\cite{mansouri2020deploying, Zhao2021}, or fog~\cite{kramer2020radar}, that adversely affect sensing in the visible or near-visible spectrum. Furthermore, LiDAR-dependent methods that are deployed for frontier exploration, as well as Search and Rescue (SAR)~\cite{matej2022sar}, are also not immune to VDCs. Therefore, it is vital to evaluate different sensors and their operational behavior in SubT environments with VDCs to enable fault tolerance by developing optimal strategies to utilize sensor fusion.

Several attempts have been made to tackle the challenges in the SubT environments for both aerial and terrestrial robotic platforms to identify and localize artifacts on the explored map~\cite{agha2021nebula, tranzatto2022cerberus, hudson2021heterogeneous, ohradzansky2021multi, li2022airdet, roucek2021system}. The development of both hardware and software solutions that are capable of addressing the related issues by utilizing multimodal perception in connection with various path-planning strategies~\cite{Zhao2021}, terrain traversability analysis~\cite{zhao2022terrain, ruetz2022travers}, and reactive navigation~\cite{mansouri2020nmpcmine} in VDCs are some of the latest trends for fully autonomous robots. Therefore, the constant demand for versatile and robust datasets from such environments with various system modalities, that are both satisfying the current requirement of state-of-the-art (SOTA) and the addition of novel sensors are the main motivations to collect the current dataset to facilitate the latest developments in this field.

Millimeter wave RADARs are a viable option for perception in SubT environments due to their resilience to VDCs~\cite{kramer2021radar, kramer2020radar, ouaknine2021carrada} as they do not require any illumination to operate~\cite{Kramer2022} and are not affected by any aerosol particles such as smoke~\cite{lu2020see} and dense vapor~\cite{kramer2021radar} that could be present in the surroundings. Moreover, due to their low power consumption and lightweight setup, they are an attractive choice to be utilized for vehicles with limited payload and power capacity such as Micro Aerial Vehicles (MAVs)~\cite{kramer2021radar}. These features are the main contributing factor that have directly increased their utilization in autonomous robots which are deployed in harsh environments~\cite{kramer2020radar, kramer2021radar, lu2020see, lu2020milliego, rapp2017probabilistic, schuster2016robust}, as well as other demanding sectors such as the automobile industry~\cite{caesar2020nuscenes, sheeny2021} and Unmanned Aerial Vehicles (UAVs) manipulation in industrial settings~\cite{Mourtzis2022}.

\begin{figure}[!t] \centering
    \includegraphics[width=1.\columnwidth]{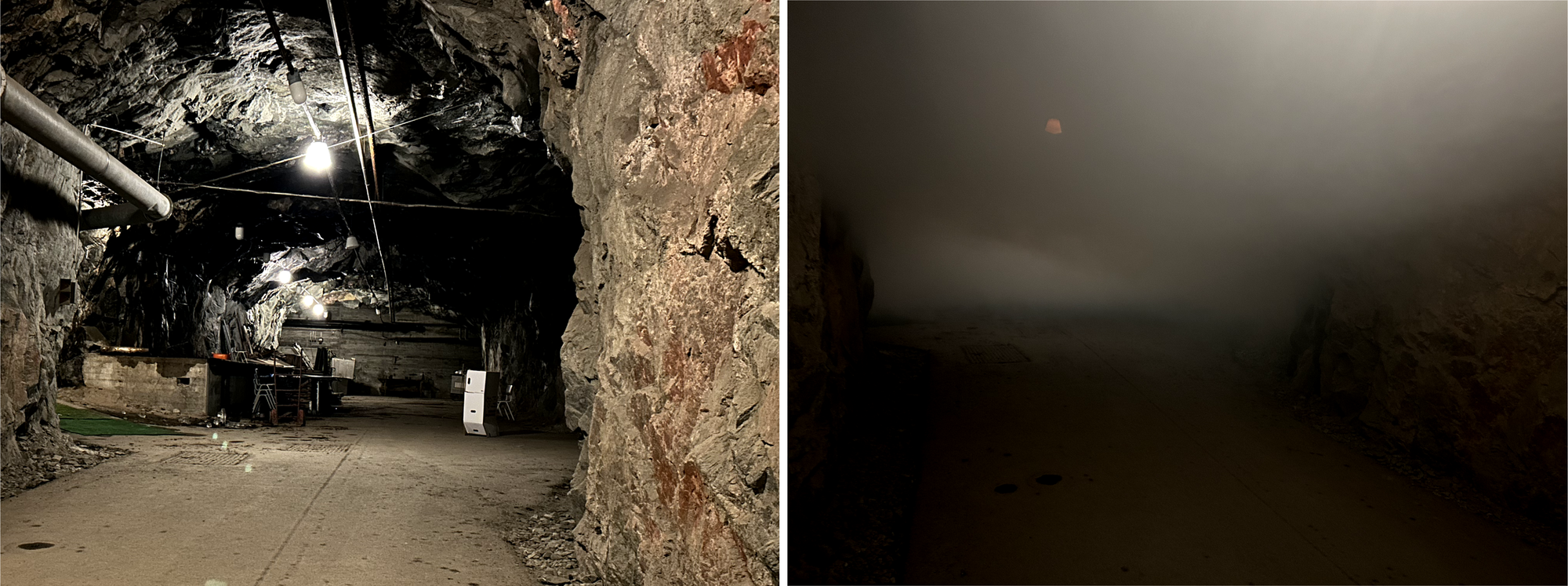}
    \setlength{\abovecaptionskip}{-10pt}
    \setlength{\belowcaptionskip}{-20pt}
    \caption{The example of SubT environment with and without dense vapor, Luleå, Sweden.}
    \label{fig:fig1_env}
\end{figure}

Given that the previously existing datasets from indoor~\cite{leung2017chilean} and under-ground environments~\cite{KOVAL2022dataset, kasper2019benchmark} are limited and do not contain measurements from a wide spectrum of available sensors in the presence of aerosol particles~\cite{Kramer2022, Zhao2021}, the main contribution of this article is to present a novel multimodal dataset that contains measurements from four RADARs that enables a 360-degree Field-of-View (FOV) of the environment, solid-state and 360-degree LiDARs, Infrared (IR)-enabled RGB-D camera and IMU in harsh SubT environment under the presence of aerosol particles, as depicted in Figure~\ref{fig:fig1_env}. 

The sensor suite is specifically designed for SubT environments, which pose unique challenges for autonomous navigation and mapping. To address these challenges, in addition to the mounted RADARs, an IR-enabled RGB-D camera and a solid-state LiDAR are deployed for the collection of the dataset. To the best of the authors' knowledge, this is the first publicly available dataset that includes both IR data with solid-state LiDAR in the presence of dense vapor in SubT environments. 

Moreover, the IR-enable RGB-D camera used in this paper allows Visual-Inertial Odometry (VIO) due to its ability to penetrate through dense vapor~\cite{kang2022IRfog} in dark environments, which is crucial for accurate object detection and collision avoidance. Additionally, we have performed a preliminary evaluation of the combined IR and RADAR data streams with potential VDCs for both static and dynamic collision avoidance. Furthermore, state estimation and mapping for the evaluation of localization from the collected dataset in the SubT environment based on Direct LiDAR Odometry (DLO)~\cite{chenk2022} is presented. Finally, the captured data are fully integrated with ROS framework~\cite{ros2009} to not only extend the previously collected dataset~\cite{KOVAL2022dataset} with VDCs but also include additional sensors such as IR-enable RGB-D camera and RADAR.

The released dataset is collected using ROS bag format to ensure its ease of usage for the research community that is focused on autonomous navigation and mapping in SubT environments as well as recording multimodal data streams with different data structures and at different rates in a time-synchronized fashion. The dataset is available for download at~\href{https://doi.org/10.5281/zenodo.7913307}{\textit{Dataset}}~\footnote{https://doi.org/10.5281/zenodo.7913307}.



\begin{figure*} [!t] \centering
     \begin{subfigure}[!t]{0.4\textwidth} \centering
         \includegraphics[width=\textwidth]{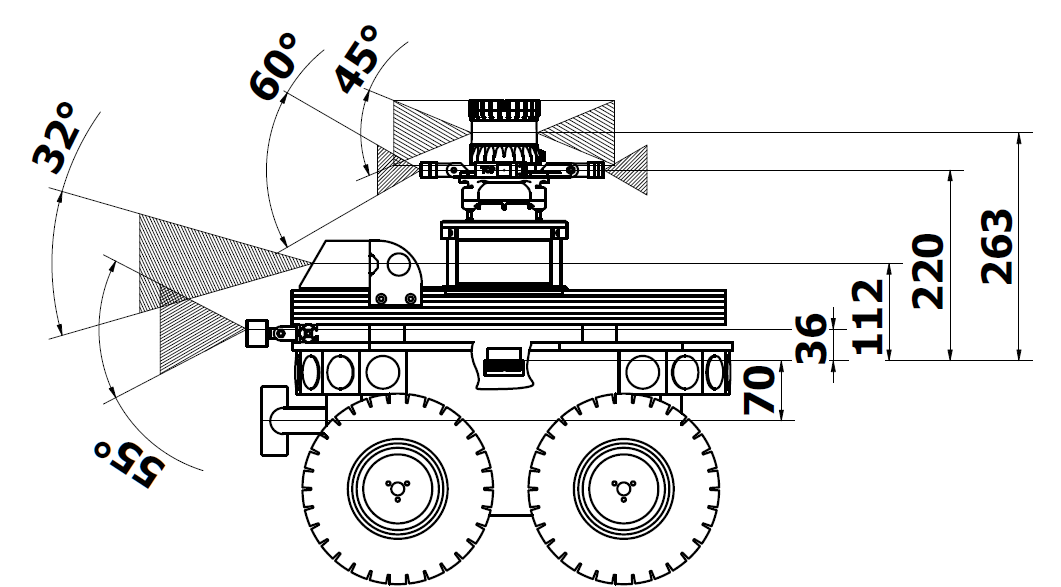}
         \caption{Model of sensor suite with corresponding FOVs.}
         \label{fig:Fig_1a}
     \end{subfigure}
     \begin{subfigure}[!t]{0.24\textwidth} \centering
         \includegraphics[width=\textwidth]{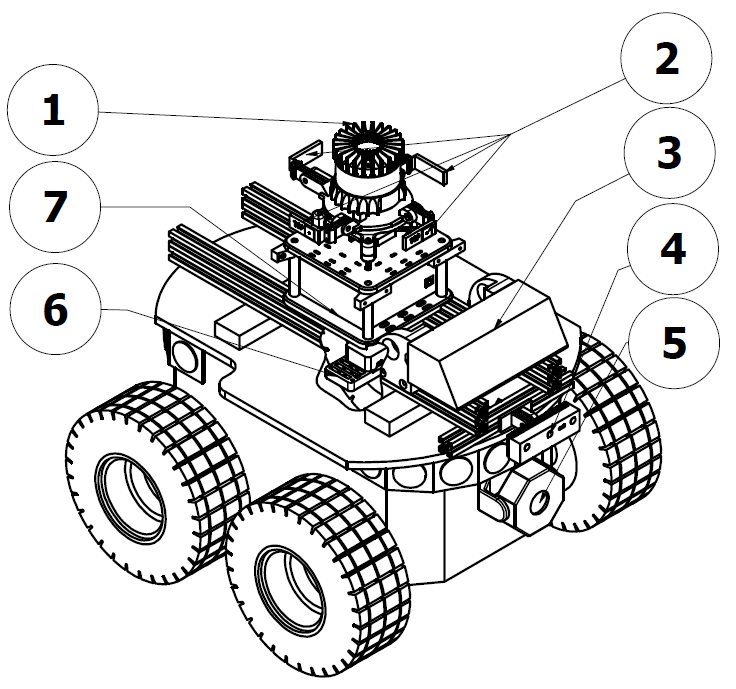}
         \caption{Detailed list of components.}
         \label{fig:Fig_1b}
     \end{subfigure}
     \begin{subfigure}[!t]{0.33\textwidth} \centering
         \includegraphics[width=0.85\textwidth]{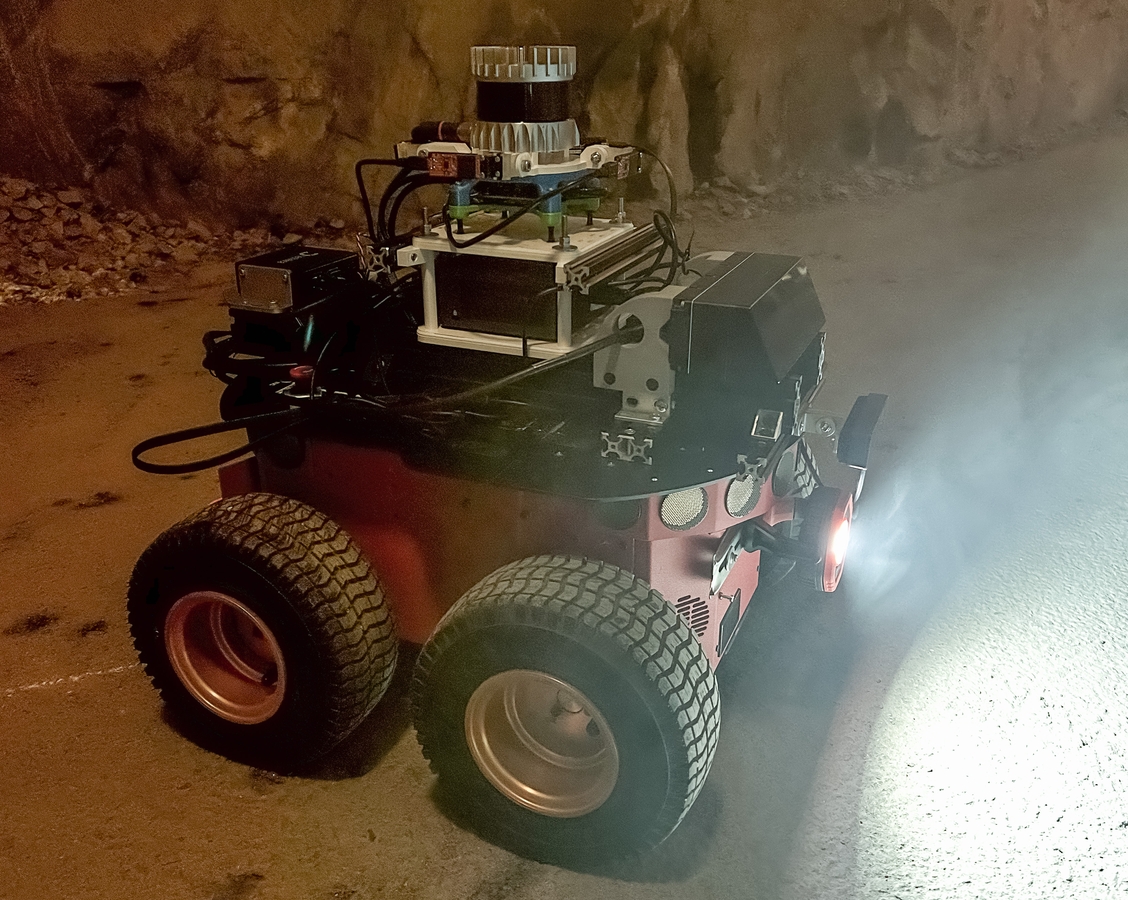}
         \caption{Pioneer with equipped sensors in SubT.}
         \label{fig:Fig_1c}
     \end{subfigure}
        \setlength{\belowcaptionskip}{-17pt}
        \caption{Data collection platform with the mounted sensor setup and detailed schematics of the location of sensors. In b) mounted components are 1-OS1-32, 2-mmWave RADARs, 3-M1600, 4-OAK-D Pro. 5-LED, 6-IMU, and 7-Intel NUC.}
        \label{fig:Fig_1}
\end{figure*}

The rest of this article is structured as follows. Section~\ref{robot} provides a brief description of the multimodal sensor suite as well as technical specifications for each sensor such as their FOVs and their expected operational behavior in the presence of VDCs. Moreover, an overview of our system architecture, its capabilities, and limitations are additionally included in section~\ref{robot}. In Section~\ref{dataset}, we present a detailed description of the data format, collection procedures, and the SubT environment where the data is collected. Furthermore, a preliminary assessment of the dataset is performed by comparing the resulting odometry based on onboard LiDARs. Moreover, the resulting map of the SubT environment is presented. Finally, we conclude this article by discussing the achieved results and future work in Section~\ref{conclusion}.

\section{Multimodal robotic platform} \label{robot}
This section will present the sensor setup and the robotic platform utilized for collecting the dataset while providing detailed technical descriptions of each sensor.

\subsection{Pioneer}

The dataset collection is achieved by utilizing a four-wheel terrestrial-based robotic platform named Pioneer 3-AT2, shown in Figure~\ref{fig:Fig_1c}, that is capable of carrying payloads of up to~\unit[30]{kg} and climbing up rough terrains with a maximum inclination of $60\%$ grade. The multimodal sensor suite, which consists of two LiDARs, four RADARs, and a stereo IR-enabled RGB-D camera, was incorporated into the platform by mounting them either directly in the front of the Pioneer or utilizing the sensor tower which is shown in Figure~\ref{fig:Fig_1b}. Figure~\ref{fig:Fig_1a} illustrates that the combined data from four mmWave RADAR modules and Ouster OS1-32 that are depicted as component 2 and 1, respectively in Figure~\ref{fig:Fig_1b} provide a full $360$-degree FOV with overlapping sections from both modalities. Moreover, the onboard \textit{Intel NUC} computer is indicated as component 7 in Figure~\ref{fig:Fig_1b}. The onboard computer unit is equipped with an Intel i7 CPU, 16 GB of RAM, and 500 GB of storage. The computing unit is powered via batteries while running Ubuntu 20.04 LTS and ROS Noetic and is placed on the lower deck of the sensor tower to facilitate its connectivity to all the onboard sensors and provide an unobstructed view of the environment for optimal data collection. To provide necessary illumination during the collection of the ground truth, a rechargeable~\unit[7]{W} LED pocket floodlight, RSPRO-WL28R~\cite{ledspec}, with a luminosity of~\unit[650]{lm} and with dimension~\unit[$93 \times 8.5 \times 30$]{mm} mounted in front of the platform below the IR-enabled RGB-D camera, OAK-D, as illustrated in Figure~\ref{fig:Fig_1a} and Figure~\ref{fig:Fig_1b}.

As depicted in Figure~\ref{fig:Fig_1}, the sensor placement on the Pioneer enables full integration and fusion of the RADAR modules with the corresponding LiDAR and IR sensor to allow navigation and object detection in VDCs. The mounted sensors and their corresponding locations on the Pioneer are shown in Figure~\ref{fig:Fig_1b}. Prior to the dataset collection, a manual and visual extrinsic calibration of all mounted sensors has been performed to synchronize the data streams from all the sensors.

\subsection{IMU}

To enable robust multimodal data fusion for mobile robots, it is essential to equip them with IMU that provides information such as orientation, linear and angular accelerations of the platform in GNSS-denied areas for instance in SubT, underwater, and space environments. Therefore, a small-size IMU unit, Pixhawk 2.1 Cube Orange, which is equipped with 3-axis accelerometers, gyroscopes, and magnetometers as well as barometric pressure sensors and has an approximate total weight of~\unit[35]{g}, is placed in the center of the platform. As illustrated in Figure~\ref{fig:Fig_1b}, the IMU unit is located below the computing unit and is indicated as component 6. The integrated accelerometer and magnetometer of the IMU module are based on LSM303D, while its gyroscope modules are based on L3GD20 and MPU9250~\cite{muhammad2021pix}.

The Pixhawk 2.1 operates at a frequency of~\unit[190]{Hz} and with its 3 instances of redundant IMU data streams provides accurate pose estimation measurements~\cite{muhammad2021pix}. It must be noted that to dampen the vibrations resulting from the motion of the platform, vibration dampers were utilized to reduce the effect of the vibration from the chassis frame on the IMU measurements for better accuracy.

\subsection{mmWave {RADAR}}

\begin{figure}[!b] \centering
   \includegraphics[width=.78\columnwidth]{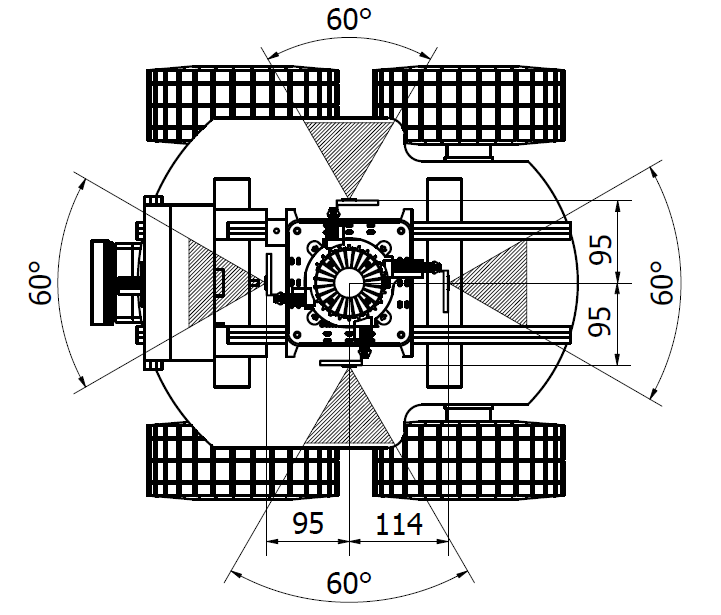}
   \setlength{\belowcaptionskip}{-5pt}
   \caption{Top view of the four mounted mmWave RADAR modules and their FOV.}
   \label{fig:fig3_radar_fov}
\end{figure}

Four mmWave RADAR modules~\footnote{https://www.mistralsolutions.com/product-engineering-services/products/som-modules/60ghz-industrial-radar-module-rom/} are used on the robotic platform. The sensors are mounted at 90$^o$ angles between each other, facing forward, left, backward, and right as illustrated in Figure~\ref{fig:fig3_radar_fov}. The RADAR modules are based on the IWR6843AoP ES2.0~\cite{radarTiModule} that are manufactured by Texas Instruments and operate at frequencies between~\unit[60]{GHz} and~\unit[64]{GHz}. Due to the capability of RADAR modules to operate in large temperature intervals, resilient to the presence of aerosol particles such as dust and smoke, as well as their small rugged form factor, these modules are suitable to be deployed in harsh SubT environments. The physical placement and the FOV of the RADARs are shown in Figure~\ref{fig:Fig_1} as component 2.

The RADAR modules are configured using the recommended configuration for "Best Range Resolution" which was suggested by the manufacturer. The resulting parameters are summarized in Table~\ref{tab:radar_parameters} and the configuration file used can be found in the ROS driver package that is provided by the manufacturer~\cite{radarROSDriver}. 

\begin{table}[!t] \centering \footnotesize
    \caption{Parameters of the RADAR modules.}
    \begin{tabular}{p{\dimexpr.5\columnwidth-2\tabcolsep-0.5\arrayrulewidth\relax} p{\dimexpr.5\columnwidth-2\tabcolsep-0.5\arrayrulewidth\relax}}
         \toprule
          RADAR & IWR6843AoP \\
         \midrule
         Frequency band &\unit[60-64]{GHz} \\
         [0.5em]
         Framerate &\unit[30]{Hz} \\
         [0.5em]
         Tx antennas & 4\\
         [0.5em]
         Rx antennas  & 3\\
         [0.5em]
         Range resolution &\unit[0.047]{m} \\
         [0.5em]
         Maximum unambiguous Range &\unit[9.02]{m}\\
         [0.5em]
         Maximum Radial Velocity &\unit[4.99]{m/s}\\
         [0.5em]
         Radial velocity resolution &\unit[0.63]{m/s}\\
         \bottomrule
    \end{tabular}
    \vspace{-7mm}
    \label{tab:radar_parameters}
\end{table}

\subsection{IR-enabled RGB-D camera}

The Pioneer is equipped with a forward-facing SOTA stereo camera with IR illumination and dot projection features to enable low-light autonomy operations. In addition, the 
camera model possesses an onboard System on Chip (SoC) to process sensor information in real-time, thereby isolating the computational needs from the onboard computer unit. OAK-D Pro~\cite{luxonis}, 
has three main built-in cameras providing RGB image output from a color camera and stereo vision, from two monocular lenses located at a baseline distance of~\unit[75]{mm}. 

The IR illumination LED has a wavelength of~\unit[930]{nm} and provides an 80$^o$ field of illumination for an input amperage of~\unit[900]{mA}. The left and right optical sensor output rectified images with a resolution of $1920 \times 720$ pixels running at a frequency of 15 frames-per-second (FPS). The dataset captured using the vision sensor contains IR-illuminated grayscale images, stereo-depth images, and the resulting depth point clouds. Additional technical specifications of the RGB-D sensor can be found in~\cite{luxonis_doc}.

\subsection{Velarray M1600}

The Velarray M1600~\cite{velarrayM1600} is a solid-state, high-performance LiDAR sensor array, designed for autonomous applications in urban, commercial, and industrial settings. With 160 channels at~\unit[10]{Hz} and wavelength of~\unit[905]{nm}, it provides a $120^o \times 32^o$ horizontal and vertical FOV with a minimum detection distance of~\unit[0.1]{m} and a maximum of~\unit[30]{m}, making it suitable for precise mapping and obstacle avoidance. The solid-state LiDAR unit with its approximate weight of~\unit[1000]{g} is placed in the front of the platform, which is indicated as component 3 in Figure~\ref{fig:Fig_1b}.

\subsection{Ouster OS1-32}

The OS1-32 LiDAR with its approximate weight of~\unit[447]{g} is commonly used for robotic applications~\cite{ousteros1}. It features 32 laser channels, with a $360^o \times 45^o$ horizontal and vertical FOV and a maximum distance of~\unit[120]{m}. The OS1-32 depicted as component 1 in Figure~\ref{fig:Fig_1b}, is mounted on top of the sensor suite above the RADAR modules for an obstructed FOV. Additionally, the sensor has a wavelength of~\unit[865]{nm} and outputs fixed-resolution image frames with depth, signal, and near-IR data at each pixel, making it suitable for semantic applications. It must be noted that both equipped LiDARs are prone to VDCs which can result in missing, noisy, or corrupted data thereby hindering the autonomous operations of the robotic platforms that are equipped with these sensors.

\section{Dataset} \label{dataset}
This section presents a comprehensive description of the environment where the data was collected and the notation and conventions used to format the acquired data. Furthermore, the data viability and usage are investigated in conjunction with robotic applications such as SLAM and collision avoidance with and without the presence of aerosol particles.

\begin{figure*} [!t] \centering
     \begin{subfigure}[!t]{0.3\textwidth} \centering
         \includegraphics[trim={1.75cm 3cm 2.6cm 4cm}, clip, width=\textwidth]{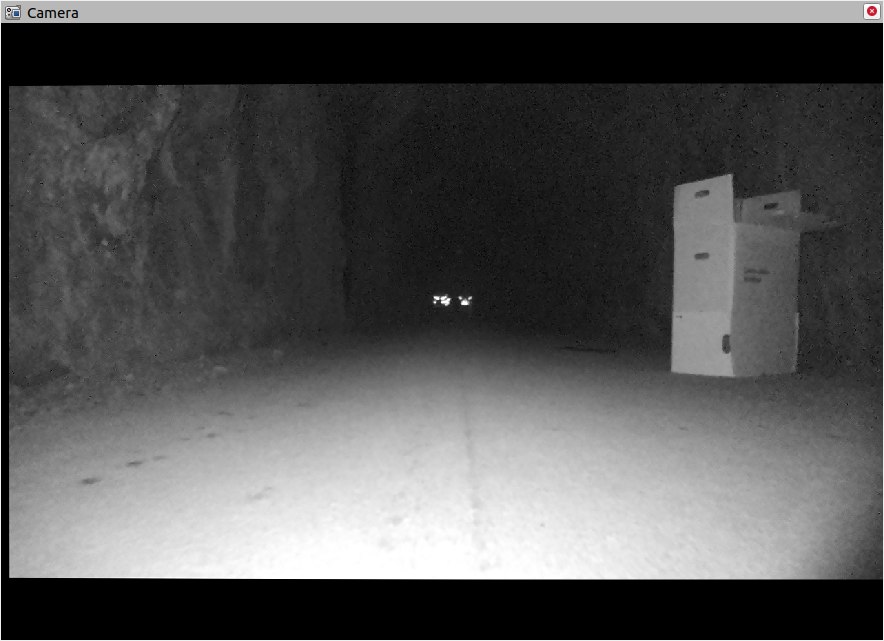}
         \caption{IR-enabled camera in Dark.}
         \label{fig:Fig_5a}
     \end{subfigure}
     \begin{subfigure}[!t]{0.31\textwidth} \centering
         \includegraphics[trim={1.75cm 1.8cm 2.6cm 4cm}, clip, width=\textwidth]{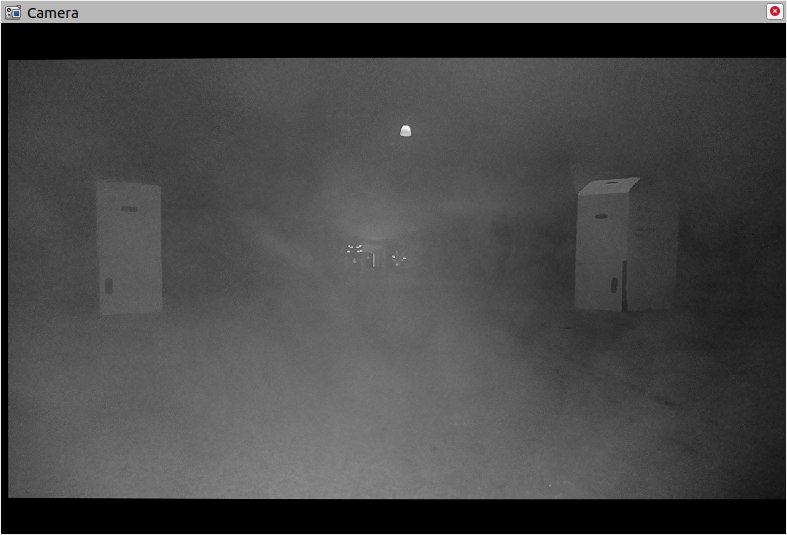}
         \caption{IR-enabled camera with dense vapor.}
         \label{fig:Fig_5b}
     \end{subfigure}
     \begin{subfigure}[!t]{0.31\textwidth} \centering
         \includegraphics[trim={1.75cm 1.8cm 2.6cm 4cm}, clip, width=\textwidth]{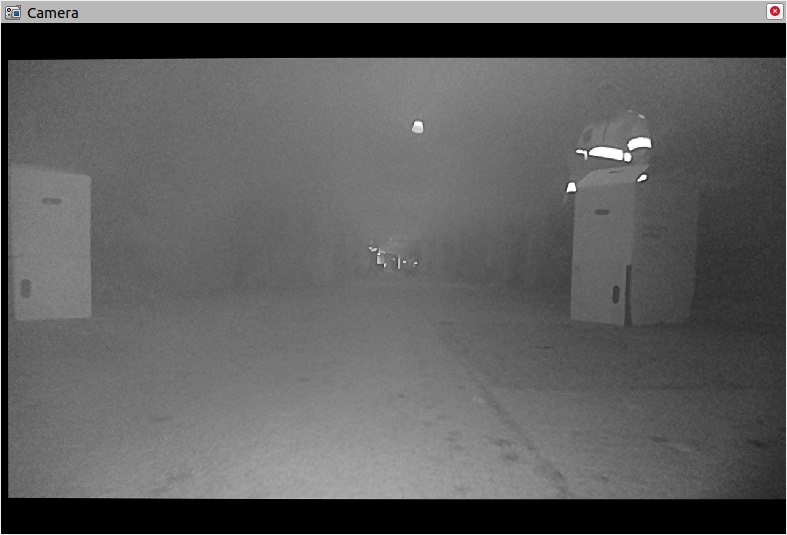}
         \caption{Detection of safety clothing with IR.}
         \label{fig:Fig_5c}
     \end{subfigure}
        \setlength{\belowcaptionskip}{-17pt}
        \caption{Evaluation of IR-illuminated camera in various VDCs for obstacle detection and SAR missions.}
        \label{fig:Fig_5}
\end{figure*}

\subsection{Environment description}

The dataset is acquired from an underground tunnel located in Luleå, Sweden. Due to challenging conditions such as poor illumination, high humidity, and additional dense vapor that is generated by a smoke machine, SLAM algorithms that solely rely on visual sensors have difficulty in creating a map of the environment, which results in localization errors that prevent autonomous navigation of the environment for the robot. Figure 1 illustrates a sample location in the SubT environments, where the data is collected. The dataset is recorded over the span of two days in three incremental batches with various VDCs and their combinations in SubT to ensure an extensive collection of operational behaviors of sensors in a variety of conditions. The Pioneer is driven with a velocity of $\sim$~\unit[1]{m/s} during the data collection by an operator to keep the platform approximately in the middle of the surrounding walls. This ensures not only the safety of the robotic platform in rough terrains but also permits the full utilization of the FOV of all mounted sensors for equal distribution of the data streams.

\begin{table}[!t]
\caption{Static transformation frames for each sensor for the collected dataset.}
\resizebox{\columnwidth}{!}{
\begin{tabular}{p{1.5cm}p{1.6cm}ll}
\toprule
Sensor&ROS Frame &Translation & Rotations\\
&& &(roll, pitch, yaw)\\
&&[meters] & [rad] \\
\midrule
Pixhawk 2.1 &base\_link& $\mathbf{p}_b$ = [0.00 0.00 0.00], & $\mathbf{r}_b$ = [0.00 0.00 0.00] \\
[0.5em]
RADAR$_{M0}$ &ti\_mmwave\_0& $\mathbf{p}_1^{pcl,1}$ = [0.01 0.00 0.00], & $\mathbf{r}_{pcl,1}$ = [0.00 0.00 0.00] \\
[0.5em]
RADAR$_{M1}$ &ti\_mmwave\_1& $\mathbf{p}_2^{pcl,2}$ = [0.00 0.01 0.00], & $\mathbf{r}_{pcl,2}$ = [0.00 0.00 1.57080]\\
[0.5em]
RADAR$_{M2}$ &ti\_mmwave\_2& $\mathbf{p}_3^{pcl,3}$ = [-0.01 0.00 0.00], & $\mathbf{r}_{pcl,3}$ = [0.00 0.00 -1.57080]\\
[0.5em]
RADAR$_{M3}$ &ti\_mmwave\_3& $\mathbf{p}_4^{pcl,4}$ = [0.00 -0.01 0.00], & $\mathbf{r}_{pcl,4}$ = [0.00 0.00 3.1416]\\
[0.5em]
RADAR$_{pcl}$ &ti\_mmwave\_pcl& $\mathbf{p}_5^{os,1}$ = [0.00 0.00 0.00], & $\mathbf{r}_{os,1}$ = [0.00 0.00 -1.57080]\\
[0.5em]
M1600 &velodyne& $\mathbf{p}_6^{b,1}$ = [0.17 0.065 0.17], & $\mathbf{r}_{b,1}$ = [3.1416 0.00 0.00] \\
[0.5em]
OS1-32 &os\_sensor& $\mathbf{p}_7^{b,2}$ = [0.00 0.0125 0.285], & $\mathbf{r}_{b,2}$ = [0.00 0.00 0.00] \\
[0.5em]
Oak-D Pro &oakd\_frame& $\mathbf{p}_8^{b,3}$ = [0.2 0.00 0.03], & $\mathbf{r}_{b,3}$ = [0.00 0.00 0.00] \\
\bottomrule
\end{tabular}}
\vspace{-6mm}
\label{tab:sensor-translation-orientation}
\end{table}

\subsection{Data format}

All the onboard sensors on the platform are mechanically fixed and the sensor suite is assumed to have a rigid body structure illustrated in Figure~\ref{fig:Fig_1b}. This allows utilization of the static transformation to extrinsically calibrate multiple sensors against the IMU frame, $b$, which is assumed as the $base\_link$ frame in the collected data. The $x$-axis of the IMU frame points towards the right direction, its $y$-axis is aligned with the forward motion of the mobile robotic platform, and the $z$-axis is gravitationally aligned. Euler angles defined as $r=[\phi,\theta,\psi]$, were utilized to define the orientation of each sensor with respect to the IMU frame and their translation noted by $p=[x, y, z]$ as shown in Table~\ref{tab:sensor-translation-orientation}. The onboard mmWave RADAR modules were noted as \textit{M0, M1, M2}, and \textit{M3}, respectively. To facilitate the usage of the dataset, they were combined into a single Point Cloud (PCL) frame, $ti\_mmwave\_pcl$, which is extrinsically calibrated with respect to OS1-32 frame, $os\_sensor$.

The timestamps for the dataset are represented in the Unix time format and all the remaining measurements are in the international system of units to enable the utilization of the dataset without any modifications. Furthermore, the timestamps of individual RADAR modules are accessible under $/ti\_mmwave/radar\_scan/$ ROS topic, where data such as intensity, velocity, and the location of the detected objects by RADAR modules are also available.

\subsection{Data usage}

To use the collected dataset, ROS provides a set of built-in functionalities called \textit{rosbag}, which provides a set of commands to access the data with ease~\cite{ros1}. To enable time synchronization, the parameter $/use\_sim\_time$ must be set as $true$ and the played bag files have the \texttt{-{}-$clock$} option activated. Moreover, the acquired data can be accessed either via individual bag files or from multiple bags by utilizing \textit{rosbag play} $sensor\_set\_id\_timestamp.bag$ and \textit{rosbag play} $sensor\_set\_id\_timestamp^*$, respectively. Additionally, the $rviz$ configuration file is provided to ease the visualization of collected data.

\begin{figure*} [!t] \centering
     \begin{subfigure}[!t]{0.46\textwidth} \centering
         \includegraphics[trim={1.75cm 8.5cm 2.6cm 10cm}, clip, width=\textwidth]{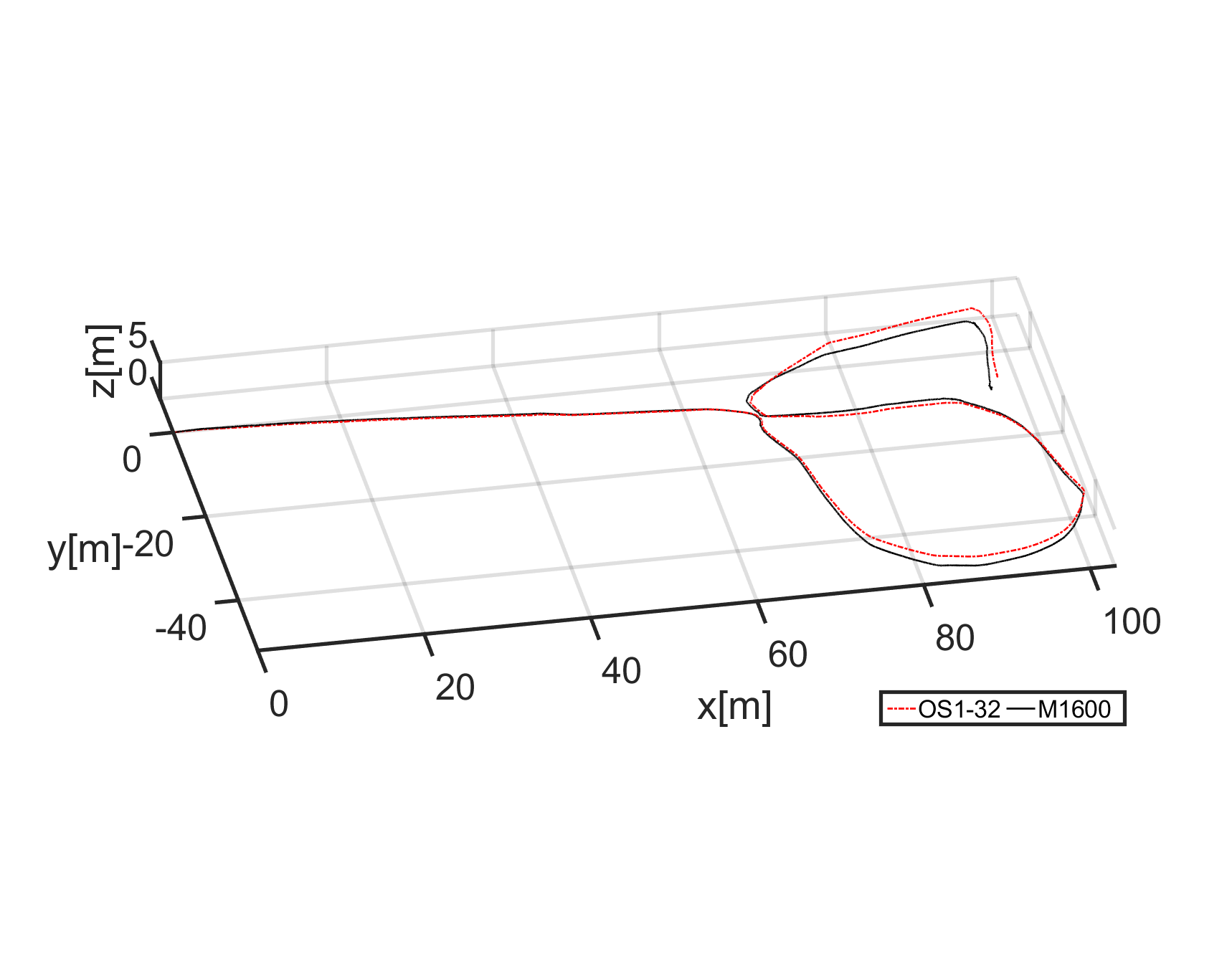}
         \caption{DLO pose estimation for both LiDARs.}
         \label{fig:Fig_6a}
     \end{subfigure}
     \begin{subfigure}[!t]{0.488\textwidth} \centering
         \includegraphics[width=\textwidth]{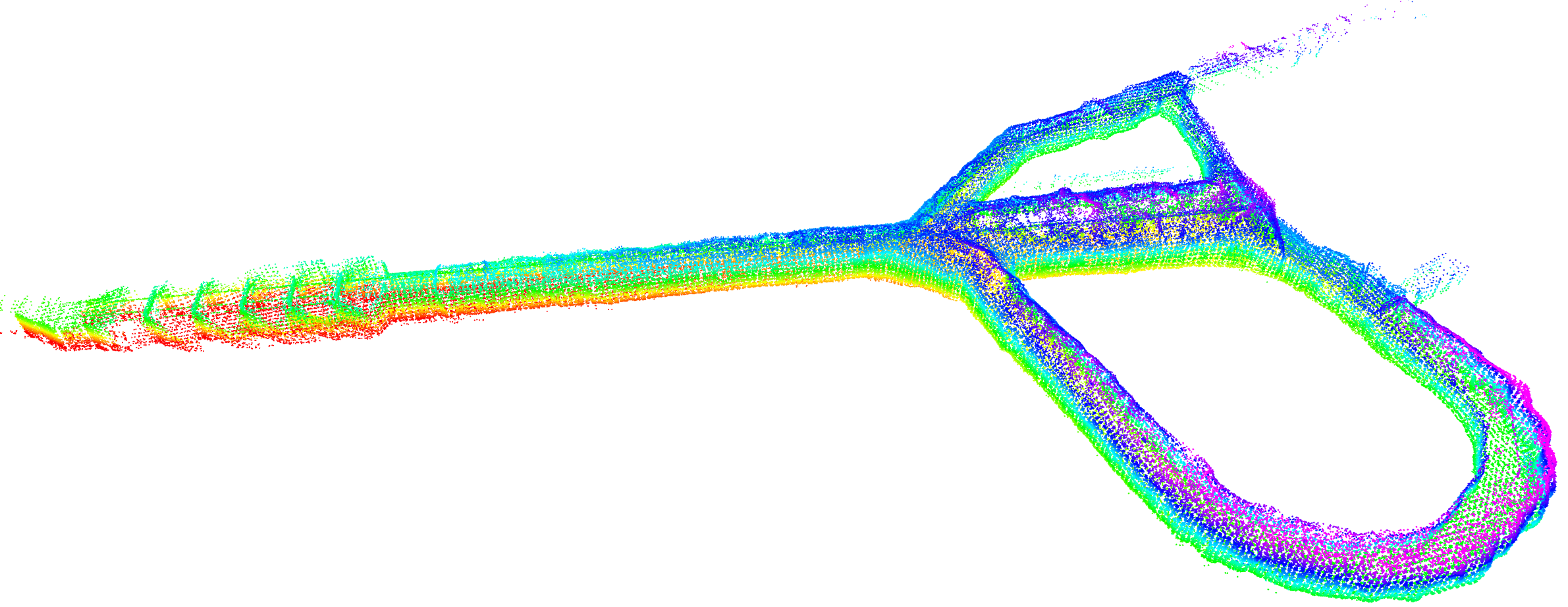}
         \caption{Generated map of SubT environment based on M1600.}
         \label{fig:Fig_6b}
     \end{subfigure}
        \setlength{\belowcaptionskip}{-20pt}
        \caption{Evaluation of trajectory estimation and mapping of the SubT environment based on the collected dataset.}
        \label{fig:Fig_6}
\end{figure*}

To evaluate the validity of the collected dataset, a preliminary assessment of the pose estimation from the onboard LiDARs based on DLO is performed. DLO~\cite{chenk2022} is a high-speed and computationally efficient frontend localization framework in which robot ego-motion is accurately resolved by utilizing the LiDAR scans and optional coupling of IMU data to achieve localization and mapping of the environment. As shown in Figure~\ref{fig:Fig_6}, DLO is used to not only generate the map of the SubT environment, where the dataset was collected but also estimate the pose of the mobile platform during its traversal based on both onboard LiDARs with high accuracy. 

To enable autonomous operations in low-light environments and in the presence of VDCs, the usage of vision sensors with active IR is investigated. As depicted in Figure~\ref{fig:Fig_5}, an IR-enabled camera, OAK-D Pro, allowed the detection of obstacles and reflective surfaces such as safety clothing in both poorly illuminated environments and in the presence of dense vapor. Thus, ensuring the viability of such sensors in deployment for SAR operations and reactive navigation schemes in SubT environments. Figure~\ref{fig:Fig_5} presents instances of dataset-captured data in various use-cases explored in this article.


\section{Conclusions} \label{conclusion}

In this paper, a multimodal dataset focused on robotic perception in harsh and unstructured SubT environments with the presence of aerosol particles is presented. The provided dataset is specifically designed to be deployed for frontier exploration studies in the presence of VDCs. Several use-case examples are provided to not only validate the dataset but also demonstrate possible future research directions where the current data can be utilized. Furthermore, a diverse range of modalities in various VDCs is provided that includes not only the compounded effect of such environments on the onboard perception systems but also isolates these effects based on individual environmental challenges such as lack of illumination and aerosol particles, thereby expediting the comprehension of the impact of VDCs on each sensor to allow the utilization of sensor fusion for further advancement in frontier exploration. 


\bibliographystyle{IEEEtran}
\bibliography{Main}


\end{document}